\documentclass[runningheads]{llncs}
\usepackage[T1]{fontenc}
\usepackage{graphicx}
\usepackage{amsmath,amssymb}
\usepackage{booktabs}
\usepackage{multirow}
\usepackage{float}
\usepackage{placeins}
\usepackage{url}
\newcommand{\ModelVP}{\text{ICI-Time}}

\DeclareMathOperator*{\argmin}{arg\,min}

\begin{document}
\title{ICI-Time: In-Context Inpainting for Adaptable Time Series Forecasting}
\titlerunning{In-Context Inpainting for Time Series Forecasting}
%
\author{Thang Nguyen
\and Dung Nguyen \and Romero Morais \and Truyen Tran}
%
\authorrunning{T. Nguyen et al.}
%
\institute{Applied Artificial Intelligence Initiative, Deakin University,\\
Geelong, Victoria 3216, Australia\\
\email{minh.t.nguyen@deakin.edu.au}}
%
\maketitle
\begin{abstract}
We propose $\ModelVP$, a novel framework that reframes time series forecasting as a visual inpainting task, leveraging the generalisation power of large vision models (LVMs). Unlike methods that require specialised temporal architectures and extensive domain-specific training, $\ModelVP$ transforms time series into structured visual representations (area charts) and applies visual in-context learning, reformulating forecasting as pattern completion within a grid-structured prompt that pre-trained vision transformers can solve without fine-tuning or architectural modification. Temporal dependencies are represented through spatial layout, with a consistent, invertible mapping between numerical and visual domains. Extensive experiments across epidemiology, meteorology, and power systems demonstrate that $\ModelVP$ performs competitively against deep learning baselines and shows promising adaptability under limited-data settings, introducing a new paradigm that bridges temporal and visual domains.

\end{abstract}
\section{Introduction}
Time series forecasting is a foundational problem in machine learning, underpinning applications across finance, epidemiology, power systems, and climate modeling. Despite decades of progress, forecasting remains inherently challenging due to complex temporal dependencies, multi-scale variability, non-stationarities, and the scarcity of labeled data in many real-world settings. Deep learning approaches have achieved state-of-the-art results \cite{liu2022pyraformer,zhou2021informer}, but they typically require domain-specific architectures and extensive training or fine-tuning when adapting to new domains or tasks. This reliance on task-specific engineering limits their scalability and hinders generalisation, especially in low-data regimes.

In contrast, foundation models in language \cite{brown2020language,wei2022chain} and vision \cite{bar2022visual} have demonstrated unprecedented generalisation through in-context learning (ICL)---the ability to solve new tasks purely through exposure to examples at inference time, without parameter updates. Recent efforts have explored ICL for temporal data using large language models (LLMs) \cite{zhou2023onefitsall} or time series models \cite{lu2025incontext}, but leveraging large vision models (LVMs) for time series forecasting remains an open challenge, primarily because temporal signals are not natively visual.

\begin{figure}[!t]
    \centering
	\includegraphics[width=0.7\columnwidth]{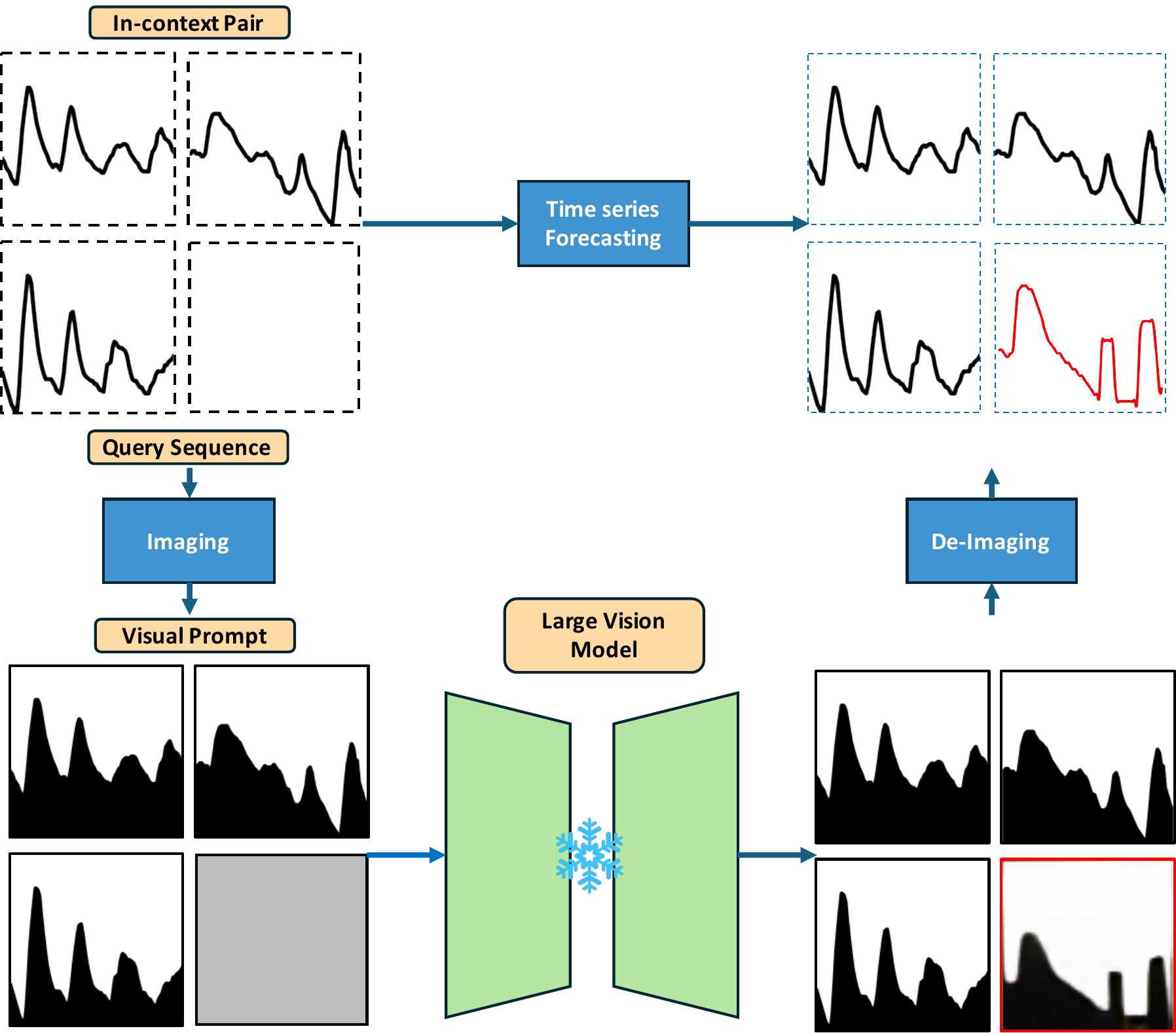}
	\caption{Visual prompting for time series forecasting via image-based in-context learning. Input sequences are converted into images and arranged in a grid: the first row holds an in-context input--output example pair, the second row holds the query with the missing forecast (gray). A pre-trained large vision model inpaints the missing region, and the prediction (red) is recovered through a de-imaging process.}
	\label{fig:Visual-prompting-via-Image-Inpainting}
\end{figure}

In this paper, we introduce $\ModelVP$ (\textbf{I}n-\textbf{C}ontext \textbf{I}npainting for \textbf{Time} series), a novel framework that redefines time series forecasting as a promptable visual inpainting task. $\ModelVP$ (i) transforms time series into visual representations (area charts), (ii) assembles input--output example pairs and the query instance into a grid-like visual prompt, (iii) employs a pre-trained LVM to forecast by completing the missing region of the image---analogous to pattern completion, and (iv) translates the generated chart back to numerical time series. As shown in Figure~\ref{fig:Visual-prompting-via-Image-Inpainting}, the LVM infers the task from the in-context example pair $(x_{1},y_{1})$ and inpaints the forecast for the query $x_{q}$ (shown in red), all without task-specific fine-tuning. Crucially, the LVM is used off-the-shelf, without any modification. This paradigm therefore (i) removes the need to train domain-specific models, (ii) supports rapid adaptation to new forecasting tasks through flexible visual prompting, and (iii) exploits the rich pattern recognition capabilities of LVMs without architectural changes or additional supervision.

Our contributions are fourfold:
\begin{itemize}
	\item \emph{A cross-modal forecasting framework} that bridges temporal and visual domains, enabling fast adaptation using off-the-shelf vision transformers.
	\item \emph{A new formulation of visual in-context learning for time series}, extending ICL beyond its NLP roots to a paradigm where temporal prediction is solved through visual prompts and inpainting.
	\item \emph{A carefully designed bidirectional mapping between time series and visual spaces} that represents temporal dependencies via spatial layout and is invertible, ensuring no loss of forecasting fidelity during transformation.
	\item \emph{Strong empirical validation} in epidemiology (ILI), meteorology (Weather), and power systems (ETT), where $\ModelVP$ matches strong Transformer-based baselines without any training and is markedly more robust in low-data regimes.
\end{itemize}

Our findings suggest that visual reasoning models can generalise to temporal tasks when equipped with suitable representations, opening a new research direction in harnessing cross-modal transfer for time series analysis.

\section{Related Work}
\paragraph{Visual In-context Learning}

In-context learning (ICL) allows a model to condition at inference time on contextual input--output examples and generate the output for a new input without parameter updates, providing a shortcut to adaptability in AI \cite{dong2024survey}. While text-based ICL emerged in autoregressive language models, \textit{Visual In-Context Learning} (VICL) trains deep networks to fill in patches of grid-like images \cite{bar2022visual}. Pioneering works such as Painter \cite{wang2023images} and SegGPT \cite{wang2023seggpt} showed that generalist vision models can perform diverse tasks---from depth estimation to semantic segmentation---by treating them as inpainting conditioned on visual examples. In the temporal domain, WeatherGFM \cite{zhao2024weathergfm} applied this paradigm to multi-modal weather data using grid-based prompts. However, general-purpose VICL for univariate time series forecasting using simple line plots remains underexplored, and key factors such as informative example selection \cite{zhang2023makes} have been studied for semantic tasks but not for temporal dynamics. Our work bridges this gap by designing a visual prompting scheme that activates the forecasting capability of LVMs without parameter updates.

\paragraph{Foundation Models and Time Series}

Foundation models have enabled new solutions for time series forecasting \cite{zhang2024large,su2024large,ye2024survey}. Early work adapted LLMs: Chronos \cite{ansari2024chronos} and Time-LLM \cite{jintime} tokenise time series via quantization or text encoding, but such methods often struggle with the ``modality gap'' between continuous numerical data and discrete text tokens \cite{ni2025harnessing}. A newer wave of ``vision-first'' models has emerged: VisionTS \cite{chen2025visionts} shows that visual Masked Autoencoders pre-trained on ImageNet can serve as zero-shot forecasters by reconstructing masked time-series images, but requires a pre-training task aligned with forecasting; ViTime \cite{yang2024vitime} trains a foundation model in a binary image metric space for robust probabilistic forecasting. This vision-first trend has since accelerated: VisionTS++ \cite{shen2025visiontspp} continually pre-trains the visual backbone on large-scale time-series corpora to narrow the modality gap, DMMV \cite{shen2025dmmv} fuses decomposition-based multi-modal views with LVMs for long-term forecasting, and OccamVTS \cite{lyu2025occamvts} and SVTime \cite{shen2025svtime} distil the predictive priors of LVM forecasters into lightweight networks. These results strengthen the evidence that visual priors transfer to temporal data, yet each still depends on continual pre-training, fine-tuning, or distillation. Unlike these approaches, which require task-aligned pre-training, architectural adaptation, or knowledge transfer into new weights, we investigate the ``free lunch'' hypothesis \cite{chen2025visionts} using standard line plots, pre-trained generalist models, and in-context visual conditioning---without task-aligned pre-training.

\paragraph{Image-based Representation for Time-series Forecasters}

Transforming time series into images bypasses the constraints of numerical sequence modeling; a recent survey \cite{ni2025harnessing} categorises such transformations into line plots, heatmaps, and spectral images. VisionTS \cite{chen2025visionts} and TimesNet \cite{wu2023timesnet} use periodicity-based heatmaps to capture long-term dependencies, but these lose the ``shape'' of the data, whereas line plots preserve the continuity and topology critical for visual pattern recognition. ViTST \cite{li2023vitst} converts irregularly sampled series into line graphs for classification, and Time-VLM \cite{zhong2025timevlm} and VLM-TSC \cite{prithyani2024feasibility} feed line plots to Vision-Language Models, showing that explicit connectivity cues outperform scatter plots or text descriptions; DePlot \cite{liu2023deplot} and ChartLlama \cite{han2023chartllama} further prove that LVMs can extract precise numerical semantics from charts. Most recently, TimeOmni-VL \cite{guan2026timeomni} unifies time-series understanding and generation within a single vision--language model and identifies low-loss bidirectional image--series conversion as a prerequisite for numerically faithful generation---independently corroborating a central design principle of our framework---but attains it through large-scale multi-task training on a purpose-built corpus. Yet most existing image-based forecasters \cite{semenoglou2023image,li2023time,sood2021visual} train a decoder from scratch or fine-tune the vision backbone, Time-VLM merely augments forecasting with a vision-language model, and unified models such as TimeOmni-VL remain training-intensive. In contrast, we reframe forecasting entirely as visual reasoning: we use a pre-trained visual prompting model \cite{bar2022visual} as-is---no fine-tuning, no external modules---and design a prompting scheme that leverages its inherent ICL ability to forecast directly from line plot grids.

\section{Preliminaries} \label{sec:prelim}
\subsection{Time Series Forecasting}
Let $\mathbf{X} \in \mathbb{R}^{C \times L}$ represent a multivariate time series, where $L$ is the total length and $C$ is the number of channels. We partition $\mathbf{X}$ into historical series $\mathbf{X}_{I} \in \mathbb{R}^{C \times T_{I}}$ and future series $\mathbf{X}_{P} \in \mathbb{R}^{C \times T_{P}}$, with $L = T_{I} + T_{P}$. $\mathbf{X}_{t}^{j}$ is the series value at the $t$-th timestep and the $j$-th channel.

The objective is to develop a predictor $f: \mathbb{R}^{C \times L_{I}} \rightarrow \mathbb{R}^{C \times L_{P}}$ that maps a lookback window of length $L_I$ to a prediction horizon of length $L_P$. Here $T_I$ and $T_P$ represent the total available data for training and evaluation, while $L_I$ and $L_P$ define the model's fixed input and output dimensions, where $T_I \gg (L_I + L_P)$.

\subsection{In-context Learning}
In-context Learning (ICL) enables a model to answer a query without task-specific training or fine-tuning: the model is given a sequence of ``similar'' input--output pairs $\{(\mathbf{x}_{i}, \mathbf{y}_{i})\}_{i=1}^{H}$ before the query $\mathbf{x}_{q}$, and produces the output $\hat{\mathbf{y}}_{q}$ directly. This contrasts with standard supervised learning, where input--output pairs are used to train or fine-tune the model.

\section{Method}
Time series forecasting traditionally requires specialised architectures and training procedures tailored to temporal data. We challenge this paradigm by leveraging the \emph{in-context learning} capabilities of Large Vision Models (LVMs) through visual prompting: time series are transformed into visual representations and the LVM forecasts via inpainting, eliminating the need for domain-specific architectures and model training or tuning, and allowing rapid adaptation to new domains even with limited data. We call this the \emph{in-context inpainting} approach to time series forecasting.

\begin{figure}[t]
	\centering
	\includegraphics[width=0.9\linewidth]{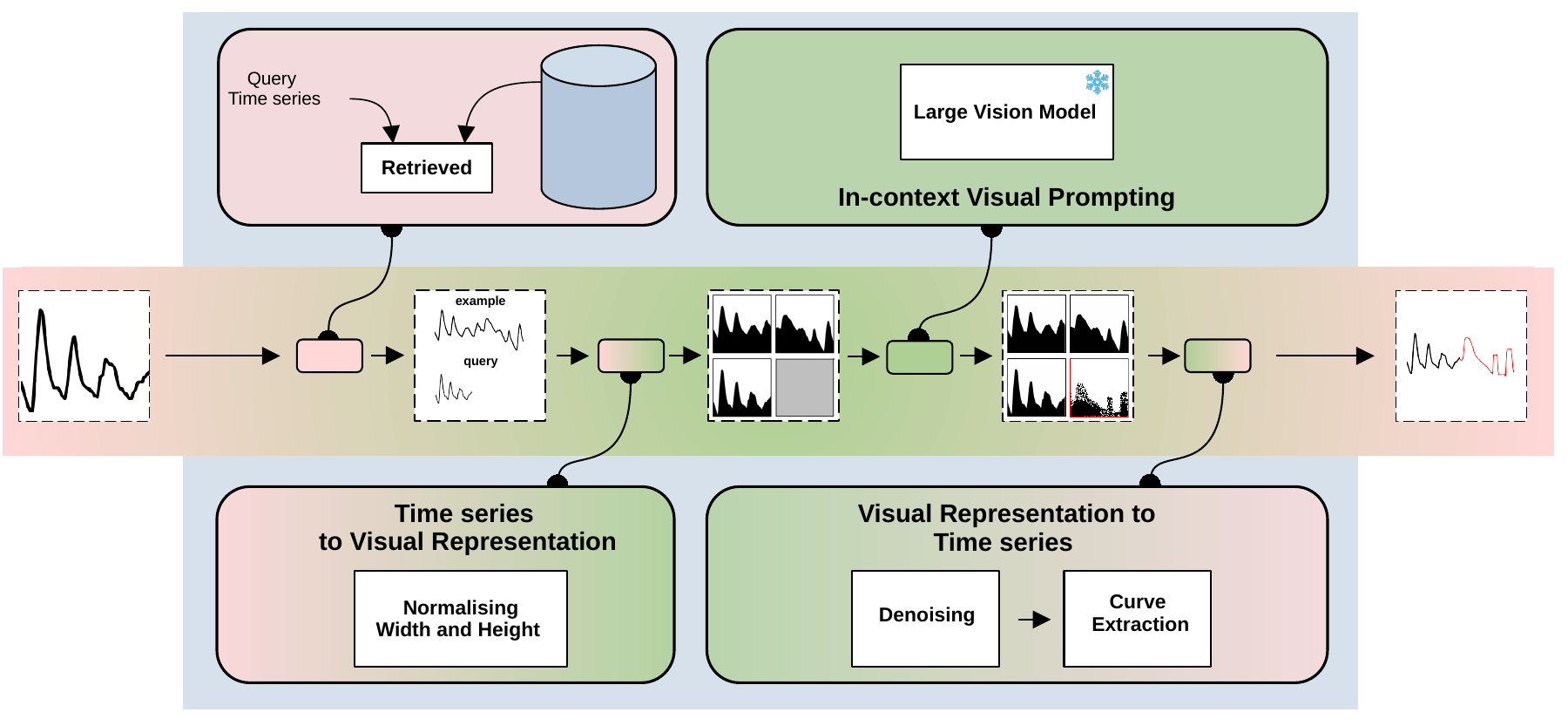}
	\caption{The ICI-Time framework: (1) example selection from historical data (top-left); (2) time-series-to-image transformation (bottom-left); (3) in-context visual prompting with a pretrained LVM \cite{bar2022visual} that inpaints the visual prompt (top-right); and (4) image-to-time-series conversion with denoising and curve extraction (bottom-right). The rightmost image shows the original series in black and the forecast in red.}
	\label{fig:framework}
\end{figure}

Our framework (Fig.~\ref{fig:framework}) comprises four components: (1) visual case-based selection from historical data, (2) conversion of time series into information-dense images, (3) an off-the-shelf LVM performing in-context inpainting, and (4) a back-conversion mechanism recovering the original time series.

\subsection{Visual Case-Based Forecasting}

Task examples enable LVMs to learn \emph{in-context}, so it is crucial to select examples relevant to the query. This resembles case-based reasoning---solving new problems by recalling and adapting similar past problems and their solutions---with the LVM acting as a nonlinear case interpolator in image space.

To enable this, we build a searchable database from $\mathbf{X}_I$ by partitioning the historical series into overlapping windows via a sliding window, each consisting of a lookback component of length $L_{I}$ and a prediction component of length $L_{P}$. For each valid starting position $s \in \{1, \dots, T_I - (L_I + L_P) + 1\}$:
\begin{equation}
	W_{s} = \left\{\mathbf{x}_{s}, \mathbf{y}_{s}\right\} = \left\{\mathbf{X}_{I}[:, s:s+L_{I}], \mathbf{X}_{I}[:, s+L_{I}:s+L_{I}+L_{P}]\right\}.
\end{equation}
This database $\mathcal{D}_{\text{hist}} = \left\{W_{s}\right\}_{s=1}^{S}$ enables pattern matching during forecasting: given the query window $W_q$, we compute the Euclidean distance between normalised inputs, $d(W_{q}, W_{s}) = ||\bar{\mathbf{x}}_{q} - \bar{\mathbf{x}}_{s}||_{2}$, and select the best example as $W_{s^{*}}=\argmin_{W_{s}} d(W_{q},W_{s})$.

\subsection{Visual Representation of Time Series} \label{sec:VP_technique}

The next step transforms numerical time series into visually interpretable formats, using normalization and boundary visualization procedures that balance information preservation and visual clarity.

\subsubsection{Image Height} \label{par:height_VP}
The image height parameter normalizes the variety of numerical scales into image sizes typically used in LVMs. Example and query windows are normalised separately.

\paragraph{Example images}
For the $i$-th example window $W_i = \{\mathbf{x}_i, \mathbf{y}_i\}$, we apply min-max scaling to both components using shared statistics:
\begin{equation}
	\bar{\mathbf{z}} \leftarrow \frac{\mathbf{z} - m_{i}}{M_{i} - m_{i}}, \qquad \mathbf{z} \in \{\mathbf{x}_{i}, \mathbf{y}_{i}\},
\end{equation}
where $m_{i} = \min(\mathbf{x}_{i}^{\text{min}}, \mathbf{y}_{i}^{\text{min}})$ and $M_{i} = \max(\mathbf{x}_{i}^{\text{max}}, \mathbf{y}_{i}^{\text{max}})$. After normalization, $\bar{\mathbf{x}}_{i}$ and $\bar{\mathbf{y}}_{i}$ are plotted into separate figures. The vertical axis extends from $h_{\text{example}}^{\text{min}} = \min(\bar{\mathbf{x}}_{i}, \bar{\mathbf{y}}_{i})$ to $h_{\text{example}}^{\text{max}} = 1.25 \times \max(\bar{\mathbf{x}}_{i}, \bar{\mathbf{y}}_{i})$. The scale $h_{\text{example}} = h_{\text{example}}^{\text{max}} - h_{\text{example}}^{\text{min}}$ serves as explicit prior knowledge transferred to the query.

\paragraph{Query images}
For a query window $W_q = \{\mathbf{x}_q, \mathbf{y}_q\}$, we normalise using the input's local statistics:
$\bar{\mathbf{x}}_{q} \leftarrow \frac{\mathbf{x}_{q} - \mathbf{x}_{q}^{\text{min}}}{\mathbf{x}_{q}^{\text{max}} - \mathbf{x}_{q}^{\text{min}}}$
and
$\bar{\mathbf{y}}_{q} \leftarrow \frac{\mathbf{y}_{q} - \mathbf{x}_{q}^{\text{min}}}{\mathbf{x}_{q}^{\text{max}} - \mathbf{x}_{q}^{\text{min}}}$
for the unknown target. The vertical axis range is $h_{\text{query}}^{\text{min}} = \min(\bar{\mathbf{x}}_{q})$ to $h_{\text{query}}^{\text{max}} = h_{\text{query}}^{\text{min}} + h_{\text{example}}$, which ensures visual consistency.

\subsubsection{Image Width} \label{par:width_VP}
Our implementation uses fixed $224 \times 224$ pixel images. To handle varying sequence lengths, we adjust the x-axis limits: $\text{xlim}_{in} = [1, L_I]$ for lookback components and $\text{xlim}_{out} = [L_I + 1, L_I + L_P]$ for prediction components.

The invertibility of our visual representation function $g: \mathbb{R}^{C \times L_P} \rightarrow \mathcal{I}$ necessitates preserving the parameters $\theta_i = \{\mathbf{x}_{i}^{\text{min}}, \mathbf{x}_{i}^{\text{max}}\}$ so that the inverse $g^{-1}: \mathcal{I} \rightarrow \mathbb{R}^{C \times L_P}$ can denormalize visual predictions via $\hat{\mathbf{y}}_{i} = \hat{\mathbf{y}}_{i}^{\text{norm}} \cdot (\mathbf{x}_{i}^{\text{max}} - \mathbf{x}_{i}^{\text{min}}) + \mathbf{x}_{i}^{\text{min}}$.

\subsection{Visual Prompting and Time Series Recovery}\label{sec:recovery}
The example and query images constitute a visual prompt; the LVM fills the empty region by producing an image $\hat{\mathbf{I}}$ via in-context learning, capturing forecasting patterns from the examples. We then recover the numerical series $\widehat{\mathbf{y}}_{q} \in \mathbb{R}^{C \times L_P}$ from $\hat{\mathbf{I}}$ in two steps.

\paragraph{Image Denoising}
We binarize, $\hat{\mathbf{I}}_{\text{binary}}(r,c) = 1$ if $\hat{\mathbf{I}}(r,c) > \tau$, apply a bitwise NOT to obtain $\hat{\mathbf{I}}_{\text{inv}}$, then apply morphological opening,
$\hat{\mathbf{I}}_{\text{opened}} = (\hat{\mathbf{I}}_{\text{inv}} \ominus \mathbf{K}) \oplus \mathbf{K}$,
where $\mathbf{K}$ is a $3 \times 3$ structuring element. The final clean image is $\hat{\mathbf{I}}_{\text{clean}} = \text{NOT}(\hat{\mathbf{I}}_{\text{opened}})$.

\paragraph{Boundary Curve Extraction}
We extract $\widehat{\mathbf{y}}_{q}$ by finding the uppermost foreground pixel in each column: for a cleaned image of height $h$, the boundary coordinate is $y_i = h - \min\{r \mid \hat{\mathbf{I}}_{\text{clean}}(r,c_i) = 1\} - 1$. To ensure continuity, we interpolate $f(c)$ such that $f(c_i) = y_i$ and sample the curve at $L_P$ equidistant points to obtain the final sequence $\widehat{\mathbf{y}}_{q}$.

\section{Experimental Results}
\subsection{Datasets}
We evaluate $\ModelVP$ across three diverse and widely benchmarked domains (summarised in Table~\ref{tab:statistics-of-datasets}):
\emph{ILI}\footnote{https://gis.cdc.gov/grasp/fluview/fluportaldashboard.html} (Influenza-Like Illness), weekly influenza-like illness patient data collected by the US CDC from 2002 to 2021, whose clear seasonal patterns and long historical record make it particularly valuable for evaluating retrieval-based strategies;
\emph{Weather}\footnote{https://www.bgc-jena.mpg.de/wetter/}, 21 meteorological indicators such as air temperature and humidity, recorded at 10-minute intervals throughout 2020;
and \emph{ETT}\footnote{https://github.com/zhouhaoyi/ETDataset} (Electricity Transformer Temperature), series from two electric transformers at 15-minute (`m') and hourly (`h') resolutions, yielding four datasets: ETTh1, ETTh2, ETTm1, and ETTm2.

\begin{table}[tbp]
	\centering
	\caption{Summary Statistics of Benchmark Datasets.}
	\label{tab:statistics-of-datasets}
	\begin{tabular}{lrrrrr}
			\toprule
			\textbf{Characteristic} & \textbf{ILI} & \textbf{Weather} & \textbf{ETTh1/2} & \textbf{ETTm1/2}\\
			\midrule
			Number of Features  & 7   & 21   & 7    & 7 \\
			Number of Timesteps & 966 & 52,696 & 17,420 & 69,680\\
			\bottomrule
	\end{tabular}
	\vspace{-0.3cm}
\end{table}

\subsection{Experimental Settings}
We adopt the data split setting from Nie et al.~\cite{nie2023time} with one key modification: we merge the training and validation data into a single database as we do not need to train or fine-tune a model, thereby enlarging our dataset. We use a look-back window $L = 96$ for our model and all Transformer-based baselines. Prediction lengths follow Nie et al.~\cite{nie2023time}, with $T \in \{24, 36, 48, 60\}$ for ILI and $T \in \{96, 192, 336, 720\}$ for the other datasets. We adopt channel independence, forecasting each variable separately, a technique proven effective in deep learning approaches \cite{nie2023time,chen2025visionts,yang2024vitime}.

We report Mean Squared Error (MSE) and Mean Absolute Error (MAE), comparing against Transformer-based baselines: Informer~\cite{zhou2021informer} (ProbSparse self-attention), Pyraformer~\cite{liu2022pyraformer} (pyramid attention), and LogTrans~\cite{li2019logtrans} (log-sparse attention). Baseline results are sourced from Nie et al.~\cite{nie2023time} when available, with additional experiments conducted to fill gaps. All experiments maintain consistent configurations to ensure fair comparison.

\subsection{Forecasting Performance}

\paragraph{Sanity check:} We implemented a nearest-neighbour baseline that simply reuses the first example from the input sequence as the prediction for all future time steps, to test whether LVMs simply copy the example over. Its results are poor compared to Transformer-based models and to $\ModelVP$, confirming that the generalisation power of LVMs comes from leveraging their vast source of visual patterns.

\subsubsection{Full-data Results}
Forecasting results are presented in Table~\ref{tab:retrieve_modelvp_performance}. $\ModelVP$ outperforms the baselines in most cases, especially on the MAE metric (23 out of 24 cases, with the remaining case being second best). This demonstrates that off-the-shelf visual in-context models can perform competitive time-series forecasting without training. The MSE metric is sensitive to noise and more reflective of the training square-loss function; in our case, noise can be suppressed by averaging over multiple runs, each using a different near-optimal prompting example.

\setlength\tabcolsep{2pt}
\begin{table}[tbp]
	\centering
	\caption{\emph{Per-horizon} performance comparison of various forecasting methods on full training data. Best results are in \textbf{bold}. The smaller the better.}
	\label{tab:retrieve_modelvp_performance}
	\renewcommand{\arraystretch}{0.9}
	\footnotesize
		\resizebox{\textwidth}{!}{%
		\begin{tabular}{l|r|cccc|cccc}
			\toprule
			Dataset & Prediction & $\ModelVP$ & Informer & Pyraformer & LogTrans & $\ModelVP$ & Informer & Pyraformer & LogTrans \\
			&  & \multicolumn{4}{c|}{MSE} & \multicolumn{4}{c}{MAE} \\
            \midrule
   			ETTh1
			& 96 	& 0.812 & 0.941 & \textbf{0.774} & 0.878 & \textbf{0.598} & 0.769 & 0.672 & 0.740 \\
			& 192 	& 1.116 & 1.007 & \textbf{0.797} & 1.037 & 0.718 & 0.786 & \textbf{0.680} & 0.824 \\
			& 336 	& 1.228 & \textbf{1.038} & 1.205 & 1.238 & \textbf{0.746} & 0.784 & 0.897 & 0.932 \\
			& 720 	& 1.291 & 1.144 & \textbf{0.977} & 1.135 & \textbf{0.757} & 0.857 & 0.788 & 0.852 \\
			\midrule
			ETTh2
			& 96 	& \textbf{0.288} & 1.549 & 0.645 & 2.116 & \textbf{0.357} & 0.952 & 0.597 & 1.197 \\
			& 192 	& \textbf{0.402} & 3.792 & 0.788 & 4.315 & \textbf{0.426} & 1.542 & 0.683 & 1.635 \\
			& 336 	& \textbf{0.449} & 4.215 & 0.907 & 1.124 & \textbf{0.455} & 1.642 & 0.747 & 1.604 \\
			& 720 	& \textbf{0.419} & 3.656 & 0.963 & 3.188 & \textbf{0.466} & 1.619 & 0.783 & 1.540 \\
			\midrule
			ETTm1
			& 96 	& 0.578 & 0.626 & \textbf{0.543} & 0.600 & \textbf{0.471} & 0.560 & 0.510 & 0.546 \\
			& 192 	& 0.673 & 0.725 & \textbf{0.557} & 0.837 & \textbf{0.527} & 0.619 & 0.537 & 0.700 \\
			& 336 	& \textbf{0.749} & 1.005 & 0.754 & 1.124 & \textbf{0.570} & 0.741 & 0.655 & 0.832 \\
			& 720 	& 1.059 & 1.133 & \textbf{0.908} & 1.153 & \textbf{0.687} & 0.845 & 0.724 & 0.820 \\
			\midrule
			ETTm2
			& 96 	& \textbf{0.198} & 0.355 & 0.435 & 0.768 & \textbf{0.288} & 0.462 & 0.507 & 0.642 \\
			& 192 	& \textbf{0.241} & 0.595 & 0.730 & 0.989 & \textbf{0.318} & 0.586 & 0.673 & 0.757 \\
			& 336 	& \textbf{0.324} & 1.270 & 1.201 & 1.334 & \textbf{0.372} & 0.871 & 0.845 & 0.872 \\
			& 720 	& \textbf{0.379} & 3.001 & 3.625 & 3.048 & \textbf{0.412} & 1.267 & 1.451 & 1.328 \\
			\midrule
			Weather
			& 96 	& \textbf{0.219} & 0.354 & 0.896 & 0.458 & \textbf{0.232} & 0.405 & 0.556 & 0.490 \\
			& 192 	& \textbf{0.280} & 0.419 & 0.622 & 0.658 & \textbf{0.273} & 0.434 & 0.624 & 0.589 \\
			& 336 	& \textbf{0.399} & 0.583 & 0.739 & 0.797 & \textbf{0.339} & 0.543 & 0.753 & 0.652 \\
			& 720 	& \textbf{0.443} & 0.916 & 1.004 & 0.869 & \textbf{0.410} & 0.705 & 0.934 & 0.675 \\
			\midrule
			ILI
			& 24 	& 3.136 & 4.657 & \textbf{1.420} & 4.480 & \textbf{1.068} & 1.449 & 2.012 & 1.444 \\
			& 36 	& \textbf{2.875} & 4.650 & 7.394 & 4.799 & \textbf{1.014} & 1.463 & 2.031 & 1.467 \\
			& 48 	& \textbf{3.572} & 5.004 & 7.551 & 4.800 & \textbf{1.098} & 1.542 & 2.057 & 1.468 \\
			& 60 	& \textbf{2.709} & 5.071 & 7.662 & 5.278 & \textbf{0.985} & 1.543 & 2.100 & 1.560 \\
			\bottomrule
		\end{tabular}}
		\vspace{-0.5cm}
\end{table}

\subsection{Analysis of Design Choices}

We ablate the key design decisions in $\ModelVP$: the height and width settings of the visual encoding, and the post-processing applied during time series recovery. Table~\ref{tab:ablation} reports results averaged over all prediction horizons; each ablation column replaces exactly one component of the full model.

\paragraph{Height Settings} The vertical axis encodes value magnitude. We compare our proposed $h_{\text{transfer}}$ (Section~\ref{par:height_VP}), which transfers the height scale from examples to queries via $h^{\text{max}}_{\text{query}} = h^{\text{min}}_{\text{query}} + h_{\text{example}}$, against $h_{1.5}$, a fixed scaling factor of 1.5 for all inputs. $h_{\text{transfer}}$ consistently outperforms $h_{1.5}$ across ETT and Weather, reducing the average MSE by 9.2\% to 18.5\% (largest on ETTh1), suggesting that maintaining visual scale consistency between examples and queries enables more effective pattern recognition. ILI is the exception, where the fixed scaling attains a lower average error.

\paragraph{Width Settings} The horizontal axis represents time. Our proposed $w_{\text{diff}}$ uses distinct temporal resolutions for input and target images ($L_I/224$ and $L_P/224$ respectively), preserving the native resolution of each component, whereas $w_{\text{even}}$ uses $L_P/224$ uniformly for both, shifting the input to the rightmost position. $w_{\text{diff}}$ outperforms $w_{\text{even}}$ on all six datasets, reducing the average MSE by 1.8\% to 5.1\% on ETT and Weather and by 10.7\% on ILI, indicating that preserving the native temporal resolution of the input consistently benefits forecasting.

\paragraph{Post-processing} A critical challenge in recovering time series from generated images is the \textit{disconnection problem}---a large forecasting error at the first prediction step, at the boundary between the input and the predicted values. We apply a Gaussian smoothing decay, weighting $w_i = \exp(-0.5 \cdot (i/\sigma)^2)$ with $\sigma = \text{window}/3$, which smoothly blends the last input point into the predictions over approximately 10 time steps. Compared with raw recovery, smoothing improves the average MSE on all six datasets, by 0.7\% to 2.2\%, with the largest gain on ILI.

\setlength\tabcolsep{2pt}
\begin{table}[tbp]
	\centering
	\caption{Ablation of design choices, \emph{averaged} over all prediction horizons. Each ablation column replaces one component of the full model ($\ModelVP$ = $h_{\text{transfer}}$ + $w_{\text{diff}}$ + smoothing). Best results are in \textbf{bold}. The smaller the better.}
	\label{tab:ablation}
	\renewcommand{\arraystretch}{0.9}
	\footnotesize
	\begin{tabular}{l|cccc|cccc}
		\toprule
		Dataset & $h_{1.5}$ & $w_{\text{even}}$ & Raw & $\ModelVP$ & $h_{1.5}$ & $w_{\text{even}}$ & Raw & $\ModelVP$ \\
		& \multicolumn{4}{c|}{MSE} & \multicolumn{4}{c}{MAE} \\
		\midrule
		ETTh1   & 1.365 & 1.151 & 1.126 & \textbf{1.112} & 0.780 & 0.719 & 0.712 & \textbf{0.705} \\
		ETTh2   & 0.446 & 0.411 & 0.394 & \textbf{0.390} & 0.462 & 0.443 & 0.430 & \textbf{0.426} \\
		ETTm1   & 0.860 & 0.795 & 0.777 & \textbf{0.765} & 0.597 & 0.573 & 0.570 & \textbf{0.564} \\
		ETTm2   & 0.316 & 0.293 & 0.288 & \textbf{0.286} & 0.368 & 0.354 & 0.351 & \textbf{0.348} \\
		Weather & 0.369 & 0.341 & 0.340 & \textbf{0.335} & 0.324 & 0.315 & 0.316 & \textbf{0.314} \\
		ILI     & \textbf{2.934} & 3.440 & 3.141 & 3.073 & \textbf{0.996} & 1.127 & 1.064 & 1.041 \\
		\bottomrule
	\end{tabular}
	\vspace{-0.3cm}
\end{table}

\subsubsection{Few-shot adaptation}

We evaluate few-shot adaptation by restricting all methods to the first $1\%$, $5\%$, or $10\%$ of the training data, following Zhou et al.~\cite{zhou2023onefitsall}; this simulates forecasting well beyond the temporal range of the training data. Informer and Pyraformer are trained under the same restricted protocol. Since $\ModelVP$ leverages historical samples directly as in-context examples, we implemented a masking procedure to prevent data leakage: when retrieving examples at test time, the initial portion of the test inputs $\mathbf{X}_I$ that would be unavailable in a real deployment is replaced with zeros, so distances are computed against zero-filled rather than actual historical values. This preserves the benefit of retrieval while maintaining the integrity of the few-shot conditions.

The results are presented in Table~\ref{tab:mse_mae_10pct_first} ($10\%$), Table~\ref{tab:mse_mae_5pct_first} ($5\%$), and Table~\ref{tab:mse_mae_1pct_first} ($1\%$), all reporting \emph{per-horizon} results; at $5\%$ and $1\%$, only the horizons providing sufficient data to train the baselines are included. $\ModelVP$ maintains robust performance across restricted data regimes, often achieving errors that are multiples lower than the baselines. For instance, on ETTh2 ($96$ pred.\ length) with $10\%$ data, $\ModelVP$ achieves an MSE of 0.317, while Informer and Pyraformer struggle at 4.047 and 4.065, respectively. At $1\%$, only ETTm1, ETTm2, and Weather provide sufficient sequence length to train the Transformer-based baselines; even in these extreme cases $\ModelVP$ dominates, e.g., on ETTm2 ($96$ pred.\ length) it maintains an MSE of 0.208, whereas Informer's error increases to 1.984.

\begin{figure}[tbp]
	\centering
	\includegraphics[width=0.6\linewidth]{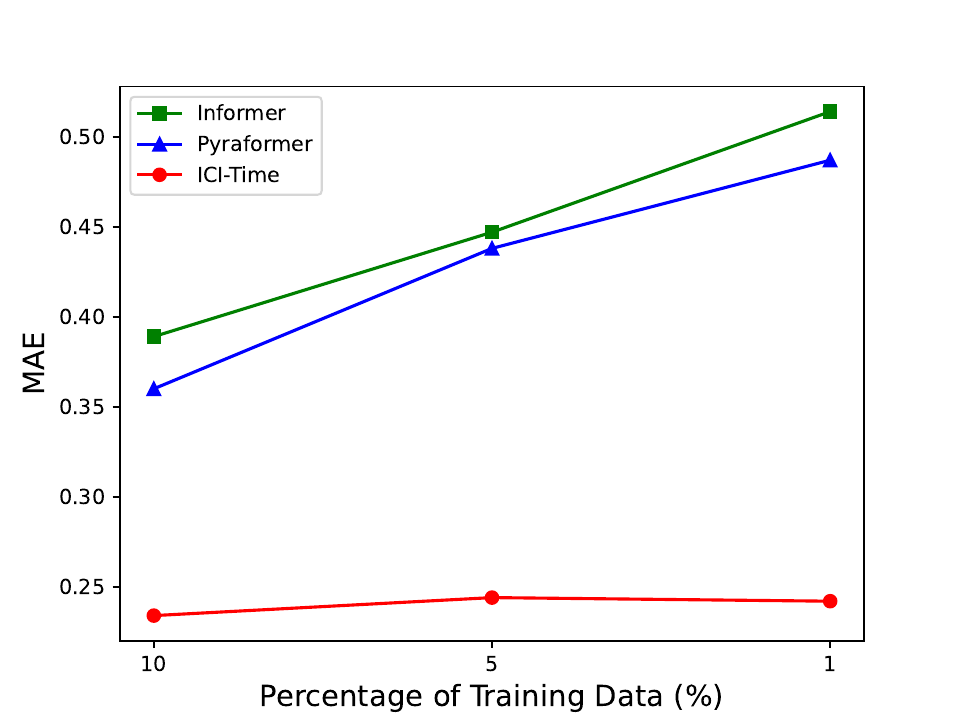}
	\caption{MAE of $\ModelVP$, Informer, and Pyraformer on the Weather dataset across percentages of training data, at prediction length $T = 96$.}
	\label{fig:weather_performance_few_shot}
\end{figure}

Figure~\ref{fig:weather_performance_few_shot} plots the MAE as a function of training data size for the Weather dataset (96 pred.\ length). As data decreases from $10\%$ to $1\%$, the MAE of Informer increases from 0.389 to 0.514 ($+32.1\%$) and Pyraformer from 0.360 to 0.487 ($+35.3\%$), whereas $\ModelVP$ remains remarkably stable, moving only from 0.234 to 0.242 ($+3.4\%$). This highlights the superior data efficiency and adaptability of $\ModelVP$ in low-resource settings.

\setlength\tabcolsep{3pt}
\begin{table}[tbp]
	\centering
	\caption{\emph{Per-horizon} few-shot performance of ICI-Time, Informer, and Pyraformer using only the first 10\% of training data. Best results are in \textbf{bold}. The smaller the better.}
	\label{tab:mse_mae_10pct_first}
	\renewcommand{\arraystretch}{0.9}
	\footnotesize
		\resizebox{\textwidth}{!}{%
\begin{tabular}{l|r|ccc|ccc}
			\toprule
			Dataset & Prediction & ICI-Time & Informer & Pyraformer & ICI-Time & Informer & Pyraformer \\
			&              & \multicolumn{3}{c|}{MSE} & \multicolumn{3}{c}{MAE} \\
			\midrule
			ETTh1
			& 96  & \textbf{1.046} & 2.113 & 1.998 & \textbf{0.687} & 1.170 & 0.984 \\
			& 192 & \textbf{1.310} & 1.913 & 1.987 & \textbf{0.793} & 1.000 & 0.978 \\
			& 336 & \textbf{1.407} & 2.099 & 2.382 & \textbf{0.834} & 1.021 & 1.053 \\
			& 720 & \textbf{1.281} & 2.332 & 1.550 & \textbf{0.800} & 1.067 & 0.909 \\
			\midrule
			ETTh2
			& 96  & \textbf{0.317} & 4.047 & 4.065 & \textbf{0.376} & 1.602 & 1.591 \\
			& 192 & \textbf{0.447} & 3.996 & 4.452 & \textbf{0.445} & 1.533 & 1.661 \\
			& 336 & \textbf{0.511} & 4.082 & 4.797 & \textbf{0.488} & 1.568 & 1.721 \\
			& 720 & \textbf{0.461} & 4.867 & 4.404 & \textbf{0.480} & 1.736 & 1.650 \\
			\midrule
			ETTm1
			& 96  & \textbf{0.743} & 1.641 & 1.570 & \textbf{0.560} & 0.937 & 0.934 \\
			& 192 & \textbf{0.888} & 2.037 & 1.726 & \textbf{0.622} & 1.104 & 0.996 \\
			& 336 & \textbf{0.971} & 2.188 & 1.998 & \textbf{0.655} & 1.126 & 1.017 \\
			& 720 & \textbf{1.140} & 2.799 & 1.920 & \textbf{0.727} & 1.273 & 1.029 \\
			\midrule
			ETTm2
			& 96  & \textbf{0.195} & 3.080 & 2.142 & \textbf{0.291} & 1.376 & 1.164 \\
			& 192 & \textbf{0.222} & 3.440 & 3.031 & \textbf{0.310} & 1.449 & 1.393 \\
			& 336 & \textbf{0.323} & 2.944 & 2.479 & \textbf{0.374} & 1.349 & 1.258 \\
			& 720 & \textbf{0.384} & 4.052 & 3.064 & \textbf{0.416} & 1.599 & 1.377 \\
			\midrule
			Weather
			& 96  & \textbf{0.232} & 0.357 & 0.288 & \textbf{0.234} & 0.389 & 0.360 \\
			& 192 & \textbf{0.292} & 0.425 & 0.364 & \textbf{0.277} & 0.427 & 0.400 \\
			& 336 & \textbf{0.398} & 0.720 & 0.465 & \textbf{0.339} & 0.534 & 0.439 \\
			& 720 & 0.478 & 0.698 & \textbf{0.466} & \textbf{0.411} & 0.528 & 0.436 \\
			\midrule
			ILI
			& 24  & \textbf{4.412} & 7.675 & 8.341 & \textbf{1.268} & 2.019 & 2.142 \\
			& 36  & \textbf{4.491} & 7.595 & 7.606 & \textbf{1.308} & 2.020 & 2.018 \\
			& 48  & \textbf{4.757} & 7.665 & 7.591 & \textbf{1.320} & 2.037 & 2.019 \\
			& 60  & \textbf{3.969} & 8.118 & 7.950 & \textbf{1.261} & 2.114 & 2.080 \\
			\bottomrule
	\end{tabular}}
\end{table}

\setlength\tabcolsep{3pt}
\begin{table}[tbp]
	\centering
	\caption{\emph{Per-horizon} few-shot performance of ICI-Time, Informer, and Pyraformer using only the first 5\% of training data; only the prediction horizons providing sufficient data to train the Transformer-based baselines are shown. Best results are in \textbf{bold}. The smaller the better.}
	\label{tab:mse_mae_5pct_first}
	\renewcommand{\arraystretch}{0.9}
	\footnotesize
		\resizebox{\textwidth}{!}{%
\begin{tabular}{l|r|ccc|ccc}
			\toprule
			Dataset & Prediction & ICI-Time & Informer & Pyraformer & ICI-Time & Informer & Pyraformer \\
			&              & \multicolumn{3}{c|}{MSE} & \multicolumn{3}{c}{MAE} \\
			\midrule
			ETTh1
			& 96  & \textbf{1.176} & 1.802 & 2.225 & \textbf{0.743} & 0.976 & 0.998 \\
			& 192 & \textbf{1.416} & 1.770 & 1.869 & \textbf{0.853} & 0.972 & 0.957 \\
			& 336 & \textbf{1.353} & 2.069 & 1.478 & \textbf{0.833} & 1.015 & 0.907 \\
			\midrule
			ETTh2
			& 96  & \textbf{0.316} & 3.720 & 4.609 & \textbf{0.375} & 1.546 & 1.706 \\
			& 192 & \textbf{0.444} & 3.884 & 4.946 & \textbf{0.448} & 1.532 & 1.758 \\
			& 336 & \textbf{0.501} & 4.319 & 3.898 & \textbf{0.482} & 1.624 & 1.549 \\
			\midrule
			ETTm1
			& 96  & \textbf{1.140} & 1.690 & 1.669 & \textbf{0.667} & 0.996 & 0.941 \\
			& 192 & \textbf{1.404} & 1.828 & 1.882 & \textbf{0.749} & 1.033 & 1.004 \\
			& 336 & \textbf{1.402} & 2.115 & 2.074 & \textbf{0.765} & 1.087 & 1.003 \\
			& 720 & \textbf{1.177} & 2.180 & 2.062 & \textbf{0.740} & 1.084 & 1.065 \\
			\midrule
			ETTm2
			& 96  & \textbf{0.209} & 2.707 & 2.197 & \textbf{0.304} & 1.304 & 1.185 \\
			& 192 & \textbf{0.232} & 3.084 & 2.567 & \textbf{0.323} & 1.398 & 1.285 \\
			& 336 & \textbf{0.337} & 2.917 & 2.548 & \textbf{0.382} & 1.365 & 1.268 \\
			& 720 & \textbf{0.382} & 3.573 & 3.368 & \textbf{0.417} & 1.502 & 1.442 \\
			\midrule
			Weather
			& 96  & \textbf{0.247} & 0.441 & 0.405 & \textbf{0.244} & 0.447 & 0.438 \\
			& 192 & \textbf{0.308} & 0.563 & 0.441 & \textbf{0.285} & 0.500 & 0.448 \\
			& 336 & \textbf{0.409} & 0.646 & 0.444 & \textbf{0.352} & 0.540 & 0.430 \\
			& 720 & 0.486 & 0.578 & \textbf{0.471} & \textbf{0.415} & 0.519 & 0.451 \\
			\midrule
			ILI
			& 24  & \textbf{4.621} & 7.453 & 7.707 & \textbf{1.312} & 2.002 & 2.046 \\
			\bottomrule
	\end{tabular}}
\end{table}

\setlength\tabcolsep{3pt}
\begin{table}[tbp]
	\centering
	\caption{\emph{Per-horizon} few-shot performance using only the first 1\% of training data; only the datasets shown provide sufficient data to train the Transformer-based baselines. Best results are in \textbf{bold}. The smaller the better.}
	\label{tab:mse_mae_1pct_first}
	\renewcommand{\arraystretch}{0.9}
	\footnotesize
		\resizebox{\textwidth}{!}{%
\begin{tabular}{l|r|ccc|ccc}
			\toprule
			Dataset & Prediction & ICI-Time & Informer & Pyraformer & ICI-Time & Informer & Pyraformer \\
			&              & \multicolumn{3}{c|}{MSE} & \multicolumn{3}{c}{MAE} \\
			\midrule
			ETTm1
			& 96  & \textbf{0.960} & 1.682 & 1.851 & \textbf{0.641} & 0.963 & 1.045 \\
			& 192 & \textbf{1.296} & 1.743 & 1.526 & \textbf{0.743} & 1.004 & 0.922 \\
			& 336 & \textbf{1.381} & 1.405 & 1.386 & \textbf{0.789} & 0.895 & 0.894 \\
			\midrule
			ETTm2
			& 96  & \textbf{0.208} & 1.984 & 2.444 & \textbf{0.308} & 1.118 & 1.264 \\
			& 192 & \textbf{0.234} & 3.631 & 3.029 & \textbf{0.324} & 1.478 & 1.402 \\
			& 336 & \textbf{0.333} & 3.340 & 3.171 & \textbf{0.380} & 1.446 & 1.404 \\
			\midrule
			Weather
			& 96  & \textbf{0.225} & 0.511 & 0.499 & \textbf{0.242} & 0.514 & 0.487 \\
			& 192 & \textbf{0.266} & 0.690 & 0.587 & \textbf{0.280} & 0.592 & 0.537 \\
			& 336 & \textbf{0.366} & 0.729 & 0.871 & \textbf{0.339} & 0.645 & 0.724 \\
			\bottomrule
	\end{tabular}}
\end{table}

\FloatBarrier

\section{Conclusion}
We have shown that visual reasoning can be successfully adapted to model complex temporal dynamics. By transforming time series into structured images and applying in-context inpainting with pre-trained vision models, ${\ModelVP}$ bypasses the need for specialised temporal architectures and costly training or fine-tuning, while achieving competitive forecasting accuracy across diverse domains. Beyond accuracy, our results reveal a broader insight: pre-trained visual models, combined with carefully designed representations, can generalise far beyond their original modalities, challenging conventional boundaries between temporal and visual modelling. Future research may explore more advanced retrieval strategies for in-context examples, and extensions to time series anomaly detection (e.g., using forecasting error for anomaly scoring) and classification (e.g., encoding classes as visual objects).

\paragraph{Data Availability.}
The datasets used in this study are publicly available on the internet at \url{https://github.com/thuml/Autoformer}.

%
%
\bibliographystyle{splncs04}
\bibliography{mybibfile}

\end{document}